# AN IMPROVED CTGAN FOR DATA PROCESSING METHOD OF IMBALANCED DISK FAILURE

Jingbo Jia[1], Peng Wu[1] and Hussain Dawood[2]

[1]School of Information Science and Engineering, University of Jinan, Jinan, China
[2]Department of Information Engineering Technology, National Skills University Islamabad, Islamabad

## ABSTRACT

*To address the problem of insufficient failure data generated by disks and the imbalance between the number of normal and failure data. The existing Conditional Tabular Generative Adversarial Networks(CTGAN) deep learning methods have been proven to be effective in solving imbalance disk failure data. But CTGAN cannot learn the internal information of disk failure data very well. In this paper, a fault diagnosis method based on improved CTGAN, a classifier for specific category discrimination is added and a discriminator generate adversarial network based on residual network is proposed. We named it Residual Conditional Tabular Generative Adversarial Networks (RCTGAN). Firstly, to enhance the stability of system a residual network is utilized. RCTGAN uses a small amount of real failure data to synthesize fake fault data; Then, the synthesized data is mixed with the real data to balance the amount of normal and failure data; Finally, four classifier (multilayer perceptron, support vector machine, decision tree, random forest) models are trained using the balanced data set, and the performance of the models is evaluated using G-mean. The experimental results show that the data synthesized by the RCTGAN can further improve the fault diagnosis accuracy of the classifier.*

## 1. INTRODUCTION

With digitization, exponential increase in information is being observed in last two decades. Based on the current increase in data, it's been predicted that around 463 Exabyte (EB) of data will be generated every day by 2025[1]. The generated data needs to be stored on disk, the stored data will lost permanently due to disk failure. Therefor disk failure prediction plays an important role to avoid the unseen circumstances. Most of the researchers[2][3][4][5][6][7] on disk failure prediction are based on SMART (Self-Monitoring Analysis and Reporting Technology) data, however the amount of failed disks data is extremely small. In data mining, it's been observed that a few classes are more important as compared to all available classes. The imbalance between the number of normal data and failure data has created hindrances. The imbalance data have generated problems in different applications such as industry[8], Internet industry[14][15][16], medical industry[9][10][11][12][13], etc.

With the continuous development of deep learning, generative Adversarial Networks (GANs)[17] is the most popular generation task algorithm today, proposed by lan Goodfellow et al. in 2014. Since it was proposed, it has achieved extremely successful applications in image and text information processing. Generative adversarial networks are currently used mainly for unstructured data such as images and text. There is a relative lack of research on the application of generative adversarial network models to structured data. Many researchers have proposed many variants on the basis of GAN to solve the problem of unbalanced amount of data categories in tabular types. For example, MedGAN[18] proposed a medical generative adversarial network to generate realistic synthetic medical records. TableGAN[19] useds generative adversarial networks to synthesize fake tables that are statistically like the original tables. Conditional Tabular

Generative Adversarial Networks(CTGAN)[20] is a GAN-based method to model tabular data distribution and sample rows from the distribution. In CTGAN, they invented the mode-specific normalization to overcome the non-Gaussian and multimodal distribution. They designed a conditional generator and training-by-sampling to deal with the imbalanced discrete columns. CTGAN model used continuous data and adds the conditional loss to discrete data to synthesize high quality data. CTGAN models continuous data and adds the conditional loss to discrete data to synthesize high quality data.

Inspired by CTGAN, improveds CTGAN is proposed to generate disk failure data.

The contributions of this paper are summarized as follows:

1. The artificial neural network of the discriminator in CTGAN is replaced with a residual network to enhance the stability of the neural network when generating adversarial networks for adversarial training.

2. Add a classifier based on the CTGAN framework. The classifier further improves the ability of the model to synthesize data by distinguishing fault samples, normal samples and generated samples.

3. The classifier loss and the discriminator loss are combined into a total loss of discriminating the true and false samples to against the generator loss.

The experimental results show that the RCTGAN algorithm can generate synthetic data with the same distribution as the original data.

The remaining article is structured as follows: "Related work" discusses the related work to solve the imbalance of disk failure data. "Proposed method" provides the details of the proposed approach. "Experiments" discusses the datasets and neural network structure. "Results and Discussion" provides insight into performance evaluation and its comparison with CTGAN approaches. Finally, in "Conclusion", the research is concluded with some future recommendations.

## 2. RELATED WORK

Commonly, researchers solve the problem of disk failure data imbalance from three perspectives: data level method [3][21][22][23][24][25][26], algorithm level method [27][28][29] and hybrid method[6][7][30].

Data level methods, also known as external methods, can be further subdivided into data sampling methods and feature selection methods, which adjust the training set of the model at the data preprocessing level. The data level approach is one of the most applied of the many methods. [21] alleviated the imbalance in the number of positive and negative samples by removing normal disk data (negative samples) to make the ratio of positive and negative samples reach 1:10. However, in the process of under-sampling the data, the important features of the data are often liminated. Due to the presence of missing salient data, it is difficult for the classifier to learn the decision boundary between positive and negative samples, resulting in reduced classifier performance. [3] over-sampled a small number of samples. Due to the high imbalance ratio of positive and negative samples, they proposed to have a high over-sampling rate to increase the number of positive samples. However, the probability of recurrence of positive samples in the oversampling process is high, which leads to poor model generalization performance. Inappropriate use of oversampling diagnostic accuracy not only does not improve, also causes an increase in computational costs. Therefore, an intelligent oversampling method to synthesize minority class samples--Synthetic Minority Over-Sampling Technique (SMOTE) [22]. It analyzes minority category samples and randomly selects samples of similar distance for interpolation to generate new minority minority samples without duplicates. It can overcome the overfitting problem generated by random oversampling to some extent. However, its time complexity is too high due to the large number of nearest neighbor operations involved; When the minority class samples contain more noise, the SMOTE algorithm will be disturbed and propagate the noise further, which affect the performance of the classifier. With the development of GAN in synthesizing tabular data[18][19][20], some researchers have used GAN to synthesize disk failure data to achieve a balanced number of samples in each category. [23] used CTGAN to synthesize disk failure data. CTGAN generates data similar to the distribution of real samples by learning the data distribution of real samples and mixes synthetic data and real data to form a training set to train the classifier. The experimental results show that CTGAN achieves better results in balancing disk failure data. [24] improve the diagnostic accuracy by integrating two GANs to generate positive samples separately and mix them in different proportions to balance the disk failure data. [25] and [26] use transfer learning to predict disk failure data, which can also alleviate the problem of insufficient failure data to some extent.

Algorithm level methods are commonly considered as cost-sensitive methods[27] and methods of integrated learning. Classifiers using cost sensitive methods assign a different weight penalty to each input training sample, and by this higher weights can be assigned to minority class samples. The importance of the minority class is increased during the training of the model, so that the classifier is biased towards the minority class and reduces the problem of low diagnostic accuracy due to the unbalanced number of class samples. [28] designed a self-encoder that can perform quadratic encoding using long and short-term memory neural networks and fully connected layers, only normal samples are used to train the classifier model and the model only learns the data distribution of normal samples, this method avoids the effect of the imbalance between the number of normal and faulty samples and improves the generalization ability of the model. [29] proposed a disk failure prediction method based on improved random forest, using the idea of ensemble learning (Bagging algorithm) to construct multiple decision trees on the training set and synthesize the classification voting results of multiple decision trees in the final prediction. It's being proved that the improved random forest algorithm has better diagnostic accuracy and robust to noisy data and outliers and can avoid overfitting problems.

Hybrid methods are a combination of data level methods and algorithm level methods. Hybrid methods are widely used due to the continuous improvement of data level methods and algorithm level methods. [30] proposed a disk failure prediction model based on the adaptive weighted Bagging-GBDT algorithm. The data level uses cluster-based stratified under-sampling to sample normal samples multiple times. At the algorithm level, multiple GBDT (Gradient Boosting Decision Tree) subclassification models with higher prediction accuracy are established by training subsets, and then the weights of each sub-model are determined adaptively. Finally, the final disk failure prediction model is integrated through weighted hard voting. [7] addressed the sample imbalance problem by increasing the weight of positive samples and indirectly decreasing the weight of negative samples and used an online disk failure prediction model based on LightGBM. It's an improvement of GBDT, which solves the problem of GBDT high computational costs. A high prediction accuracy is achieved with a guaranteed low false alarm rate. [6] proposed pre-failure reset window as the main data processing method. This method can solve sample imbalance within a certain range, reduce potential fuzzy samples, and enhance data availability. In order to make full use of the temporal and spatial characteristics of disk data. The article also proposes a CNN-LSTM disk failure prediction method based on the combination of convolutional neural network and long short-term memory network. CNN extracts the spatial features of the data, LSTM effectively captures the dependencies between time series, and the combined model further improves the failure prediction rate of the prediction model.

CTGAN on imbalanced disk failure data demonstrated higher diagnostic accuracy using the augmented dataset compared to the original dataset[23]. Although this method can learn the data distribution of the real data, the experimental results show that there is a certain gap between the data distribution of CTGAN synthetic data and the real data distribution.

# 3. PROPOSED METHOD

## 3.1. Discriminator Network Structure

The principle of the residual network is to add the input with output of the neural network unit and activate it. In proposed, residual network concatenates the input with output of the neural network unit and then activates it. Figure 1 shows, the dimension of the input discriminator and output residual block are 710 and 256, respectively. Which are concatenated and used for next residual block.

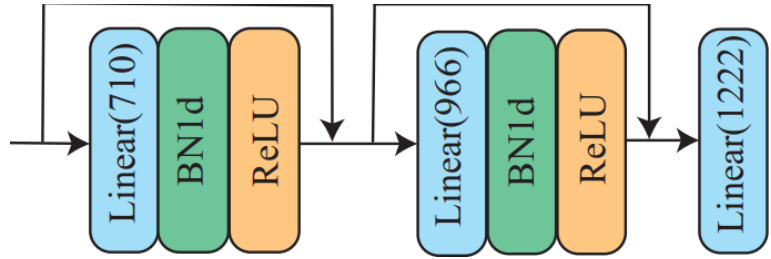

Figure 1. Residual network structure of the discriminator

We replace the original artificial neural network with residual network. The residual network is characterized by an easy optimization. The network degradation is avoided by utilizing the skip connections technique for internal residual blocks, which can recognize the identity mapping of the network and retain the original input features while increasing the number of layers, it also overcomes the problem of information loss, ensure the integrity of information, and helps to improve the generalization ability of the neural network.

## 3.2. Classifier Network Structure

This experiment adds a classifier for specific category judgment of disk data based on the CTGAN model framework. The discriminator in the CTGAN framework can only judge the sample's 'true' and 'false'. The classifier not only distinguishes between the two categories of 'true' and 'false', also determines whether the data comes from the generator. Consider there are N category samples in the real data, then the output of the classifier is N+1 categories. If the generator can 'cheat' the classifier and discriminator at the same time, it can be considered that the accuracy of the generator is improved. The classifier network is similar to the general artificial neural network as shown in Figure 2.

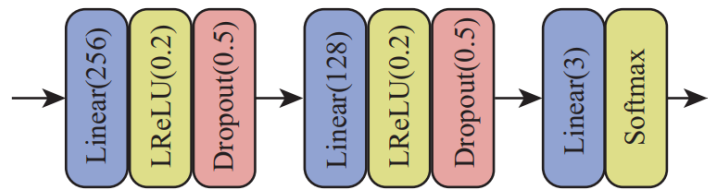

Figure 2. Network structure of the classifier

The hidden layer is a two-layer network as shown in above figure with number of neurons follow the structure from large to small, 256 and 128, respectively. Dropout layer is used to alleviate the overfitting problem. Whereas the loss rate of neurons is considered as 0.5.

## 3.3. RCTGAN Framework

The RCTGAN comprises of a generator and a discriminator. The RCTGAN model combines the PacGAN [31] algorithm and the WGAN-GP [32] algorithm. The WGAN-GP algorithm solves the problem of extreme distribution of WGAN parameters. A gradient penalty term is added to the loss function of the discriminator to replace the parameter interval limit of the discriminator in WGAN. For RCTGAN, the essence of its training is to minimize the Wasserstein distance between the real data distribution and the generated data distribution. It can make the training of RCTGAN more stable and achieve higher quality generation results. The PacGAN algorithm is introduced in the discriminator. As, it packages multiple samples into a single sample which is fed into the neural network for 'true' and 'false' sample discriminations. Although this method can alleviate the problem of model collapse, effectively. However the category information in each sample is not effectively utilized. To overcome aforementioned problem, with the basic idea of the original GAN, a classifier is added to the framework. The generator is responsible for generating data to do data augmentation. The discriminator is responsible for distinguishing the generated data from the real data, while guiding the generator to generate realistic data. The classifier distinguishes the different categories while guiding the generator to generate data for each category. This is more beneficial for classification tasks, especially for imbalanced data. A discriminator and generator structure allows the generator to be trained in more stringent way, sufficiently. In this way the salient information can be fully utilized. In PacGAN, multiple fault samples and normal samples considered as one sample, therefor the discriminator cannot distinguish among normal or faultly sample in specific category, and thus cannot share the weight between the discriminator and the classifier. Figure 3 shows the structure of the RCTGAN.

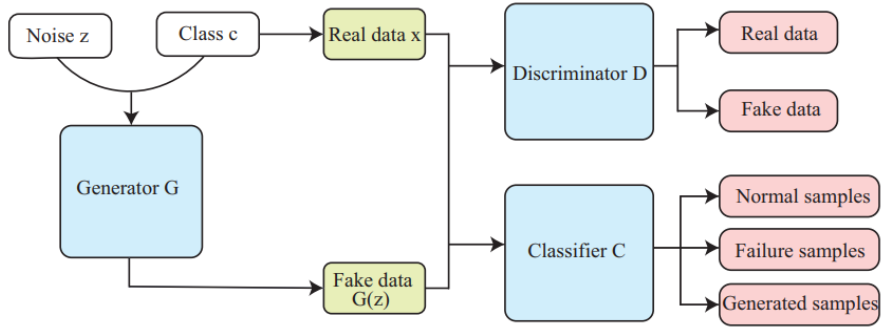

Figure 3. The illustration of our proposed RCTGAN model

When real data enters the classifier network, it determines the specific category of each sample, and guides the classifier to determine whether it is a faultly or a normal sample, correctly. While synthetic data fed into classifier network, the classifier is directed to discriminate this type as synthetic samples. Classifier is trained samples couple of times, now the classifier can distinguish specific categories of samples.

### 3.4. Comprehensive Loss Functions

For the generator, only it is not necessary to minimize the loss from the discriminator to determine 'true' and 'false', also to minimize the loss from the classifier to determine the specific category of each sample. The loss from the discriminator can be expressed as in equation 1:

$$Loss_g = -\frac{1}{n}\sum_{i=1}^{n} y\_fake_i \qquad (1)$$

where, $y\_fake_i$ represents the output result of the discriminator for the i-th sample. Where n is the total number of samples. $Loss_g$ shows an average value of the discriminator output, which is also used for adversarial training of the discriminator and the generator.

The classifier loss uses the multi-class cross entropy Loss function. The cross entropy describes the distance between the probability distributions of the real and generated samples. In other words, the smaller the cross entropy value, the closer the two probability distributions are. As shown in equation 2:

$$Loss_c = -\sum_{i=1}^{n} y\_real_i \cdot \log(y\_fake_i) \qquad (2)$$

where, the probability distribution $y\_real_i$ is the expected output and the probability distribution $y\_fake_i$ is the actual output.

The discriminator loss is the difference between the mean value of the discriminator discriminating the real data output and the mean value of the discriminating the generated data output shown in equation 3.

$$Loss_d = -\left(\frac{1}{n}\sum_{i=1}^{n} y\_real_i - \frac{1}{n}\sum_{i=1}^{n} y\_fake_i\right) \qquad (3)$$

where, y_real and y_fake represent the output values of real data and synthetic data through the discriminator, respectively. The smaller the difference, the closer the distribution of synthetic data and real data is.

The combination of the discriminator loss $Loss_d$ and the cross entropy loss of the classifier $Loss_c$ constitutes the total loss $Loss_{total}$ of the generative adversarial network to discriminate the truth of the sample, as shown in equation 4. Adversarial training with generator loss $Loss_g$ and $Loss_{total}$.

$$Loss_{total} = Loss_d + Loss_c \qquad (4)$$

## 4. EXPERIMENTS

The effectiveness of the RCTGAN over the CTGAN on synthetic data is presented. To validate the synthetic data, four classifiers are used that includes a well-performing Multi-Layer Perceptron (MLP), and three classical classification algorithms namely as Decision Tree (DT), Random Forest (RF) and Support Vector Machine (SVM). We have also done the testing on original datasets without using any data augmentation technique. To overcome the biasness, experiments are conducted multiple time and an average is considered. The result and discussion section are further sub divided into experimental setup, evaluation matrix and framework and finally the results and discussions.

### 4.1. Datasets

A publicly available data for disk inspection was considered for evaluation by Backblaze[33] in 2020. It have comprehensive data information containing the daily SMART information and operating status of the disks. The calculation formula or threshold values of SMART attribute values from different venders and models may be inconsistent. To eliminate the influence of venders and model. In experiments, the disks of the same vender and model was selected with the large number of targeted cluster. The selected disk vender is 'SEAGATE' and the disk model is 'ST4000DM000'. The proportion of imbalanced positive and negative samples set in this experiment are 1:100 and 1:500. A total of 218 failure samples were selected, and another 21800 or 109000 normal samples were selected to form the total data set. The specific data set division is shown in Table 1. It is worth noting that the training set data in the table is used for GAN model training to synthesize the dataset used.

Table 1. Description of Datasets.

| ratio | Dataset | Failure samples | Normal samples |
|---|---|---|---|
| 1:100 | Training set | 174 | 17400 |
| | Test set | 44 | 4400 |
| 1:500 | Training set | 174 | 87000 |
| | Test set | 44 | 22000 |

**4.2. Environment**

The experimental were conducted on 64-bit Windows 10, Intel Core i5-10500 processor with 8G RAM. The Python version used is 3.8 and PyTorch 1.10 was used for deep learning framework.

**4.3. Experimental Setup**

The parameter settings such as total neurons per hidden layer, optimizer, loss functions, activation function, normalization layer and learning rate are shown in Table 2. The generator and the discriminator have the same optimizer, normalization layer and learning rate except for the number of neurons in the hidden layer and loss function. The RCTGAN Classifier network is similar to the generator network with major difference of first a linear layer followed by a leakyReLu layer with alpha=0.2. The network structure of the RCTGAN model is shown in Table 3. In RCTGAN training, we use CrossEntropyLoss loss function with training data batchsize of 500 and initial learning rate of 0.0002 with Adam optimser.

Table 2. Parameter Settings.

| Parameter | Generator | Discriminator | Classifier |
|---|---|---|---|
| Total neurons per hidden layer | 386, 642 | 966, 1222 | 256, 128 |
| Optimizer | Adam | Adam | Adam |
| Loss Function | Wasserstein loss | Wasserstein loss | CrossEntropyLoss |
| Activation | ReLU | ReLU | LeakyReLU(0.2) |
| Normalization | BatchNorm1d | BatchNorm1d | - |
| Learning Rate | 0.0002 | 0.0002 | 0.0002 |

Table 3. RCTGAN Model Network Structure.

| Type | Neural network for each layer | Output dim |
|---|---|---|
| Generator G | Input | (None,130) |
| | Linear, BN, ReLU | (None,386) |
| | Linear, BN, ReLU | (None,642) |
| | Linear | (None,69) |
| Discriminator D | Input | (None,710) |
| | Linear, BN, ReLU | (None,966) |
| | Linear, BN, ReLU | (None,1222) |
| | Linear | (None,1) |
| Classifier C | Input | (None,710) |
| | Linear, LeakyReLU, Dropout | (None,256) |
| | Linear, LeakyReLU, Dropout | (None,128) |
| | Linear, Softmax | (None,3) |

## 5. RESULTS AND DISCUSSIONS

## 5.1. Evaluation Metrics

The disk failure samples are considered as a positive samples, denoted as P (Positive). The normal disk samples are called negative samples, denoted as N (Negative). For example, if a faulty sample is correctly predicted by the classifier as a faulty sample, it is considered as True positive (TP). If it is classified as a normal sample, incorrectly, it is considered as false negative (FN). Therefore, the classification results of the model have the following four possibilities. The confusion matrix is shown in Table 4:

Table 4. Confusion Matrix.

| Category | Predicted as failure | Predicted as normal |
|---|---|---|
| Actual failure | True positive(TP) | False negatives(FN) |
| Actual normal | False positive(FP) | True negatives(TN) |

Geometric mean is used for evaluation. measure the overall performance of the classifier. It is used to measure the overall performance of the classifier. The result of the geometric mean is higher only when the detection rate of both normal and fault samples is high. The G-mean can be calculated as in equation 5:

$$\text{G-mean} = \sqrt{\frac{TP}{TP+FN} \cdot \frac{TN}{TN+FP}} \quad (5)$$

## 5.2. Evaluation Framework

The dataset is divided into training set and test set with the ratio of 8:2. The training set is used to train the CTGAN and RCTGAN models to synthesize the data. The fault sample data synthesized by the model is mixed with the ratio 1:1 with the real training sample data set. The mixed data is used to train four classifiers. Finally, the diagnostic accuracy of each classifier is evaluated with the test data. The corresponding G-mean values of each classifier trained on the mixed data are compared separately. The evaluation process is shown in Figure 4.

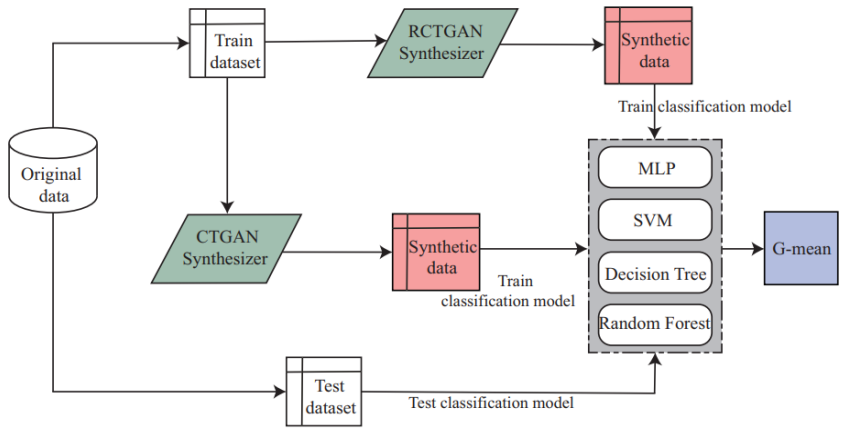

Figure 4. Evaluation framework on synthetic data and real data

## 5.3. Comparison of Training Curve

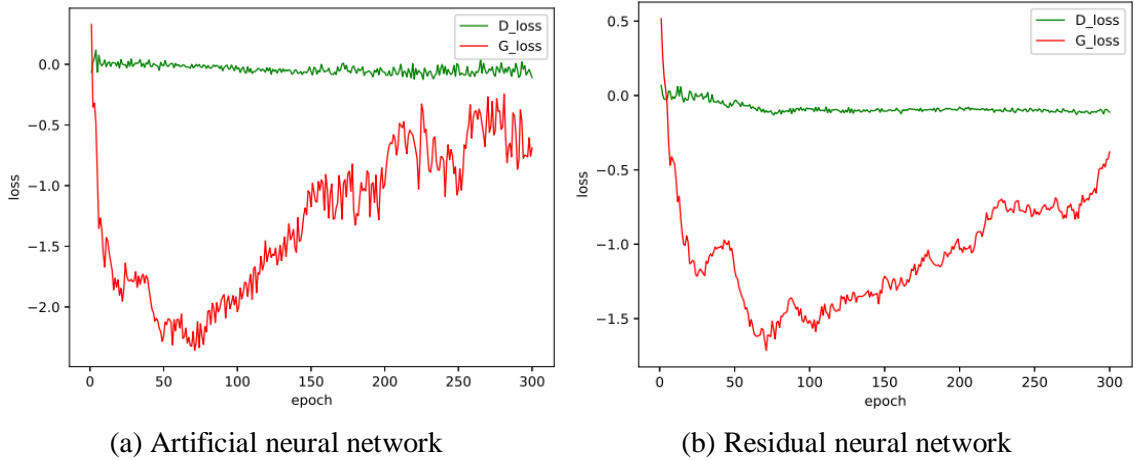

(a) Artificial neural network  (b) Residual neural network

Figure 5. Only for the discriminator improvement in CTGAN. Adversarial training plots of Artificial neural network and Residual neural network

Figure 5(a) and Figure 5(b) shows the training loss plots of the CTGAN model discriminator when using artificial neural network and residual network, respectively. The loss fluctuation of the discriminator and generator after utilizing the discriminator network with the residual network is obviously more stable, which indicate the possibility of losing data information in training process of the artificial neural network. The residual network has the advantage of jump connection and can learn the information that the artificial neural network cannot learn. The discriminator using the residual network can enhance the stability of the neural network during the adversarial training of the generative adversarial networks.

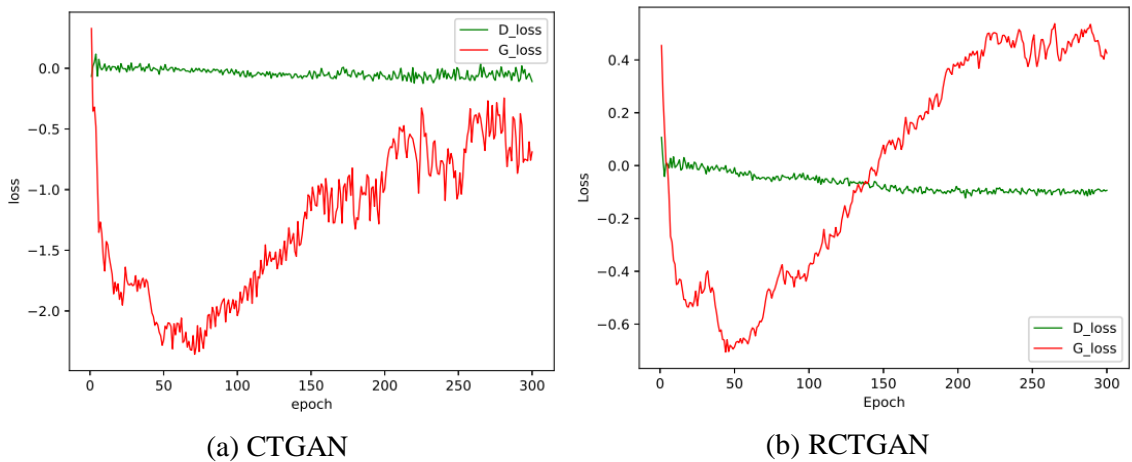

(a) CTGAN  (b) RCTGAN

Figure 6. CTGAN and RCTGAN adversarial training curve

Figure 6 shows the curves of the loss function of the generator and discriminator of the CTGAN and RCTGAN model during the training process. The discriminators and generators are trained alternately, i.e., the discriminators are trained once and then the generators are trained once. The figure shows that the discriminator is very stable, indicating that the discriminator is able to effectively adjust the generator while learning useful information from the data. It is clear the loss of discriminator D has been jittering around 0, indicating that the ability of the generator to synthesize data is also becoming stronger with the improvement of the discriminator's discrimination ability. From figure 6(b), in the initial stage of training, most of the samples generated by the generator are based on noise information, so the quality of the generated samples is low, and the samples are easily identified by the discriminator. At this time, the value of the discriminator and the generator loss function highly fluctuates. As the training times reaches at 50, the discriminator has less fluctuationes and the generator loss value increases gradually. It indicates that the generator network learns the samples features of the samples, gradually. In the late stages of training, the discriminator and generator loss values does not have considerable fluctuations. As the results of several experiments, it proves that the generator network can generate samples close to the real data.

## 5.4. Comparison of Data Distribution

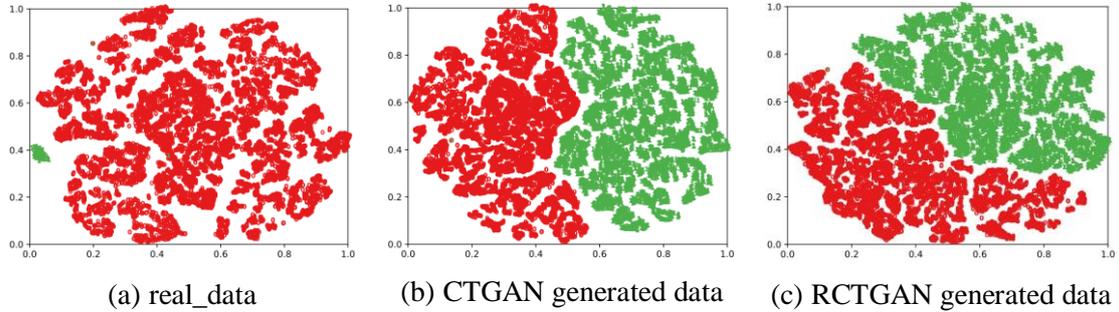

(a) real_data    (b) CTGAN generated data    (c) RCTGAN generated data

Figure 7. The red "0" represents the data distribution of negative samples, while the green "1" represents the data distribution of positive samples. The data distribution of positive and negative samples. (a) is the actual data sample. (b) and (c) represent the data distributions after augment the dataset using the generative models CTGAN and RCTGAN, respectively

It can be seen from Figure 7(a) that the distribution of positive samples is concentrated and there is a clear boundary between positive and negative samples. As shown in Figure 7(b) and 7(c), CTGAN and RCTGAN both generate positive and negative sample data with clear classification boundaries. Although the data generated by CTGAN is more scattered, there are still characteristics of small-scale clustering. In other words, it has a problem of insufficient diversity. The data generated by RCTGAN has a wide and concentrated distribution, and is more diverse.

## 5.5. Comparison of Models

In order to verify the effectiveness of the proposed method. We use DT, RF, SVM and MLP classification models to compare the quality of fake data generated by CTGAN and RCTGAN models. Each classifier uses the same hyperparameters to ensure that the results are fair. DT uses the default parameters provided by the sklearn library. After verification, the number of decision trees in RF is 45. In SVM, the penalty coefficient $C$ of the objective function is 100, the coefficient gamma of the kernel function is 1, and the kernel function is RBF. We choose the MLP with a three layers network structure and the learning rate is 0.0002.

In experiments, the results of each synthesis method on each classifier were recorded. The diagnostic accuracy of each classifier in Table 5, 6 is averaged over multiple experiments. "no augmentation" indicates that the performance of the model without any augmentation method being considered. "CTGAN" and "RCTGAN" indicates that the use of augmented models to balance the original dataset.

Table 5. The data ratio of positive and negative samples in the experiment is 1:100.

| Model | G-mean(%) | | | |
|---|---|---|---|---|
| | DT | RF | SVM | MLP |
| no augmentation | 73.31 | 62.53 | 36.92 | - |
| CTGAN | 78.04 | 74.85 | 72.01 | 87.28 |
| RCTGAN | 89.52 | 78.42 | 77.70 | 89.68 |

Without any data augmentation, we cannot build a neural network model due to the imbalance of positive and negative samples. Its clear from Table 5, the diagnostic accuracy of the data synthesized by the RCTGAN model is higher than that of the CTGAN synthesized data on each classifier model. In particular, the performance on decision trees improved by 11.5%. The improvement is 3.6% and 5.7% on random forest and Support vector machine, respectively, and only 2.4% on multi-layer perceptron.

Table 6. The data ratio of positive and negative samples in the experiment is 1:500.

| Model | G-mean(%) | | | |
|---|---|---|---|---|
| | DT | RF | SVM | MLP |
| no augmentation | 60.29 | 44.16 | 42.64 | - |
| CTGAN | 57.74 | 49.00 | 60.19 | 78.63 |
| RCTGAN | 69.21 | 54.60 | 67.17 | 80.5 |

To further demonstrate the effectiveness of RCTGAN. As shown in Table 6, we expanded the imbalance ratio of the original data. It is not possible to train an efficient classification model on the original dataset without considering data augmentation. After using CTGAN for data augmentation on the original data, the fault diagnosis accuracy of the classifier has improved. However, our improved RCTGAN model has higher diagnostic accuracy than the existing CTGAN model. When the ratio of imbalanced data is 1:500, the MLP model performs the best, with a diagnostic accuracy of 80.5%. There is also a significant improvement in performance on other models. It can be shown that our improvement can improve the generalization ability of the classifier model.

## 6. CONCLUSIONS

This paper mainly improves CTGAN based on two aspects. The first is the stability of neural networks during adversarial training based on GANs. The other is based on GAN in terms of the quality of the synthesized data. The experiments show that the improved RCTGAN model in this experiment is effective in terms of stability and diagnostic accuracy and further improves the diagnostic accuracy of the classification model. As future work, our focus is on using the hybridization of RCTGAN and Synthetic Minority Over-Sampling Technique (SMOTE) to address class imbalanced of disk failure problems.

## ACKNOWLEDGEMENTS

The research work in this paper was supported by the Shandong Provincial Natural Science Foundation of China (Grant No. ZR2019LZH003), Science and Technology Plan Project of University of Jinan (No. XKY2078) and Teaching Research Project of University of Jinan (No. J2158). Peng Wu is the author to whom all correspondence should be addressed.